%% file: 0-main.tex
\documentclass{midl}

\usepackage{bm, booktabs, siunitx, multirow, hyperref, xspace, graphics, wrapfig, caption}

\graphicspath{{./figures/}}
\newcommand{\B}{\fontseries{b}\selectfont}
\newcommand{\model}{DBGDGM\xspace}

\makeatletter 
\g@addto@macro{\@algocf@init}{\SetKwInOut{Parameter}{Hyperparameters}} 
\makeatother
\SetKwInput{kwInit}{Initialize}

\let\emptyset\varnothing

\jmlrvolume{-- Under Review}
\jmlryear{2023}
\jmlrworkshop{Full Paper -- MIDL 2023 submission}
\editors{Under Review for MIDL 2023}
\title[\model: Dynamic Brain Graph Deep Generative Model]{\model: Dynamic Brain Graph Deep Generative Model}

\midlauthor{\Name{Alexander Campbell\midljointauthortext{Contributed equally}\nametag{$^{1, 2}$}} \Email{ajrc4@cl.cam.ac.uk}\\
\Name{Simeon Spasov\nametag{$^{*1}$}} \Email{ses88@cl.cam.ac.uk} \\
\Name{Nicola Toschi\nametag{$^{1}$}}  \\
\Name{Pietro Li\`o\nametag{$^{3, 4}$}} \\
\addr $^{1}$ Department of Computer Science and Technology, University of Cambridge, United Kingdom \\
\addr $^{2}$ The Alan Turing Institute, United Kingdom \\
\addr $^{3}$ University of Rome Tor Vergata, Italy \\
\addr $^{4}$ A.A. Martinos Center for Biomedical Imaging, Harvard Medical School, United States}

\begin{document}
\maketitle
\input{1-abstract}

\input{2-introduction}
\input{4-method}
\input{5-experiments}

\input{6-conclusion}
\input{7-acknowledgments}
\bibliography{8-bibliography}
\input{9-appendix.tex}
\end{document}

%% file: 1-abstract.tex
\begin{abstract}
Graphs are a natural representation of brain activity derived from functional magnetic imaging (fMRI) data. It is well known that clusters of anatomical brain regions, known as functional connectivity networks (FCNs), encode temporal relationships which can serve as useful biomarkers for understanding brain function and dysfunction. Previous works, however, ignore the temporal dynamics of the brain and focus on static graphs. In this paper, we propose a dynamic brain graph deep generative model (DBGDGM) which simultaneously clusters brain regions into temporally evolving communities and learns dynamic unsupervised node embeddings. Specifically, \model represents brain graph nodes as embeddings sampled from a distribution over communities that evolve over time. We parameterise this community distribution using neural networks that learn from subject and node embeddings as well as past community assignments. Experiments demonstrate \model outperforms baselines in graph generation, dynamic link prediction, and is comparable for graph classification. Finally, an analysis of the learnt community distributions reveals overlap with known FCNs reported in neuroscience literature.
\end{abstract}

\begin{keywords}
Dynamic graph, generative model, functional magnetic resonance imaging
\end{keywords}

 




%% file: 2-introduction.tex
\section{Introduction}


Functional magnetic resonance imaging (fMRI) is a non-invasive imaging technique primarily used to measure blood-oxygen level dependent (BOLD) signal in the brain~\cite{huettel2004functional}. A natural representation of fMRI data is as a discrete-time graph, henceforth referred to as a dynamic brain graph (DBG), consisting of a set of fixed nodes corresponding to anatomically separated brain regions and a set of time-varying edges determined by a measure of dynamic functional connectivity (dFC)~\citep{calhoun2014chronnectome}. DBGs have been widely used in graph-based network analysis for understanding brain function~\citep{hirsch2022graph, raz2016functional} and dysfunction~\citep{alonso2020dynamics, dautricourt2022dynamic, qingbao2015dynamic}.
\clearpage \newpage

Recently, there is growing interest in using deep learning-based methods for learning representations of graph-structured data~\citep{goyal2018graph, hamilton2020graph}. A graph representation typically consists of a low-dimensional vector embedding of either the entire graph~\citep{narayanan2017graph2vec} or a part of it's structure such as nodes~\citep{grover2016node2vec}, edges~\citep{gao2019edge2vec}, or sub-graphs~\citep{adhikari2017distributed}. Although originally formulated for static graphs (i.e. not time-varying), several existing methods have been extended~\citep{mahdavi2018dynnode2vec, goyal2020dyngraph2vec}, and new ones proposed~\citep{zhou2018dynamic, sankar2020dysat}, for dynamic graphs. The embeddings are usually learnt in either a supervised or unsupervised fashion and typically used in tasks such as node classification~\citep{pareja2020evolvegcn} and dynamic link prediction~\citep{goyal2018dyngem}.


To date, very few deep learning-based methods have been designed for, or existing methods applied to, representation learning
 of DBGs. Those that do, tend to use graph neural networks (GNNs) that are designed for learning node- and graph-level embeddings for use in graph classification~\citep{kim2021learning, dahan2021improving}. Although node/graph-level embeddings are effective at representing local/global graph structure, they are less adept at representing topological structures in-between these two extremes such a clusters of nodes or communities~\citep{wang2017community}. Recent methods that explicitly incorporate community embeddings alongside node embeddings have shown improved performance for static graph representation learning tasks~\citep{sun2019vgraph, cavallari2017learning}. How to leverage the relatedness of graph, node, and community embeddings in
a unified framework for DBG representation learning remains under-explored. We refer to Appendix~\ref{appendix_related_work} for a summary of related work.

\paragraph{Contributions} To address these shortcomings, we propose \model, a hierarchical deep generative model (DGM) designed for unsupervised representing learning on DBGs derived from multi-subject fMRI data. Specifically, \model represents nodes as embeddings sampled from a distribution over communities that evolve over time. The community distribution is parameterized using neural networks (NNs) that learn from graph
and node embeddings as well as past community assignments. We evaluate \model on multiple real-world fMRI datasets and show that
it outperforms state-of-the-art baselines for graph reconstruction, dynamic link prediction, and achieves comparable results for graph classification.  

%% file: 4-method.tex
\section{Problem formulation}
\label{problem_formulation}

We consider a dataset of multi-subject DBGs derived from fMRI data $\mathcal{D} \equiv \mathcal{G}^{(1:S,\, 1:T)}=\{\mathcal{G}^{(s,\, t)}\}_{s,\, t=1}^{S,\, T}$ that share a common set of nodes $\mathcal{V}=\{v_1,\, \dots,\, v_N \}$ over $T \in \mathbb{N}$ timepoints for $S \in \mathbb{N}$ subjects. Each $\mathcal{G}^{(s,\, t)} \in \mathcal{G}^{(1:S,\, 1:T)}$ denotes a non-attributed, unweighted, and undirected brain graph snapshot for the $s$-th subject at the $t$-th timepoint. We define a brain graph snapshot as a tuple $\mathcal{G}^{(s,\, t)} = (\mathcal{V}, \mathcal{E}^{(s,\, t)})$ where $\mathcal{E}^{(s,\, t)} \subseteq \mathcal{V} \times \mathcal{V}$ denotes an edge set. The $i$-th edge for the $s$-th subject at the $t$-th timepoint $e^{(s,\, t)}_i \in  \mathcal{E}^{(s,\, t)}$ is defined $e^{(s,\, t)}_i=(w^{(s, t)}_i,\, c^{(s, t)}_i)$ where $w^{(s, t)}_i$ is a source node and $c^{(s, t)}_i$ is a target node. We assume each node corresponds to a brain region making the number of nodes $|\mathcal{V}|=V \in \mathbb{N}$ fixed over subjects and time. We also assume edges correspond to a measure of dFC allowing the number of edges $|\mathcal{E}^{(s,\, t)}| = E^{(s,\, t)} \in \mathbb{N}$ vary over subjects and time. We further assume there exists $K \in \mathbb{N}$ clusters of nodes, or communities, the membership of which dynamically changes over time for each subject. Let $z^{(s,\, t)}_i \in [1:K]$ denote the latent community assignment of the $i$-th edge for the $s$-th subject at the $t$-th timepoint. For each subject's DBG our aim is to learn, in an unsupervised fashion, graph $\boldsymbol{\alpha}^{(s)} \in \mathbb{R}^{H_{\alpha}}$, node $\boldsymbol{\phi}_{1:N}^{(s,\, t)}=[ \boldsymbol{\phi}_{n}^{(s,\, t)}] \in \mathbb{R}^{N \times H_{\phi}}$, and community $\boldsymbol{\psi}_{1:K}^{(s,\, t)} = [\boldsymbol{\psi}_{k}^{(s,\, t)}] \in \mathbb{R}^{K \times H_{\psi}}$ representations of dimensions $H_\alpha$, $H_\phi$, $H_\psi \in \mathbb{N}$, respectively, for use in a variety of downstream tasks. 

\section{Method}
\label{method}

\begin{wrapfigure}{r}{0.5\textwidth}
  \centering
    \includegraphics[width=0.5\textwidth, draft=false]{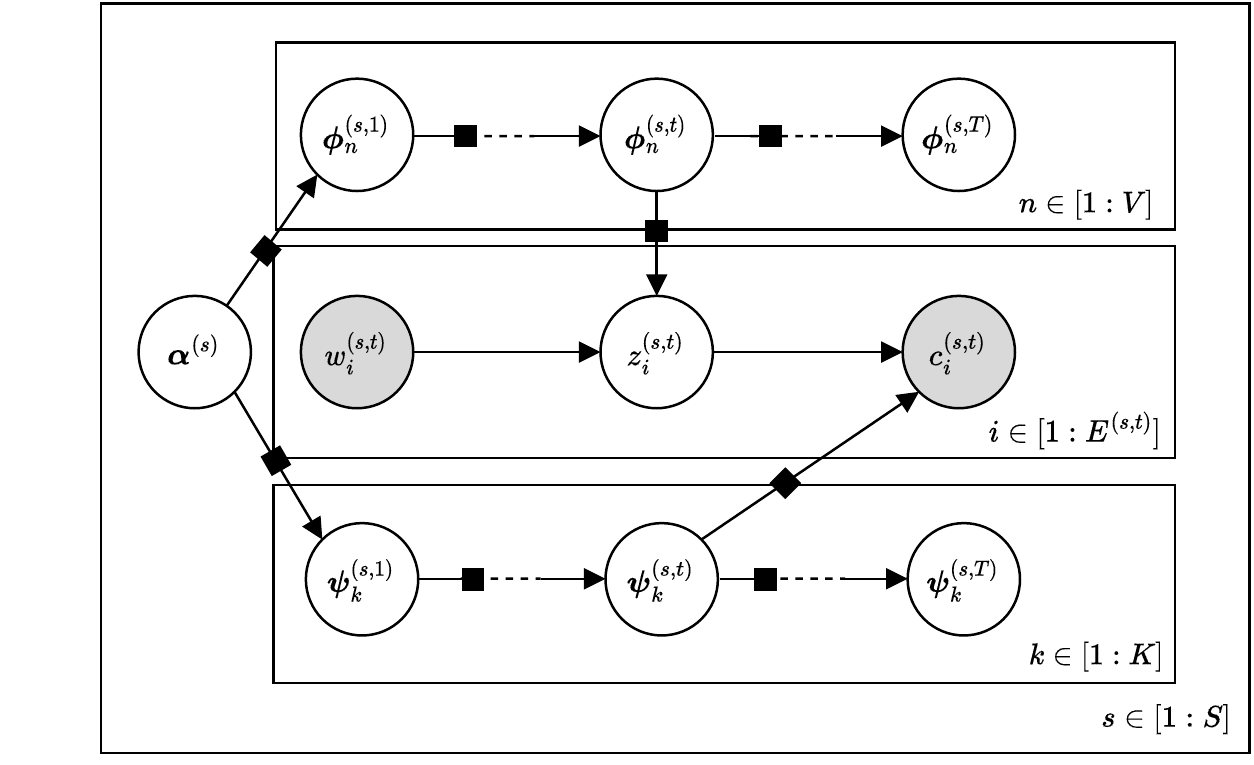}
  \caption{Plate diagram for \model. Latent and observed variables are denoted by white-and gray-shaded circles, respectively. Solid black squares denote non-linear mappings parameterized by NNs.}
  \label{fig:model}
\end{wrapfigure}

\model defines a hierarchical deep generative model and inference network for the end-to-end learning of graph, node, and community embeddings from multi-subject DBG data. Specifically, \model treats the embeddings and edge community assignments as latent random variables collectively denoted $\Omega^{(s,\, t)} = \{\boldsymbol{\alpha}^{(s)},\,$ $\boldsymbol{\phi}^{(s,\, t)}_{1:N},\,$ $\boldsymbol{\psi}^{(s,\, t)}_{1:K}, \{z_i^{(s,\, t)}\}_{i=1}^{E^{(s,\, t)}}\}$, which along with the observed DBGs, defines a probabilistic latent variable model with joint density $p_\theta(\mathcal{G}^{1:S,\,1:T}, \Omega^{1:S,\, 1:T})$.


\subsection{Generative model}
\label{generative_model}

\paragraph{Graph embeddings} We begin the generative process by sampling graph embeddings from a prior $\boldsymbol{\alpha}^{(s)} \sim p_{\theta_\alpha}(\boldsymbol{\alpha}^{(s)})$ implemented as a normal distribution following
\begin{equation}
p_{\theta_\alpha}(\boldsymbol{\alpha}^{(s)}) = \text{Normal}(\boldsymbol{0}_{H_{\alpha}},\, \mathbf{I}_{H_{\alpha}})
\label{eq:alpha_prior}
\end{equation}
where $\boldsymbol{0}_{H_{\alpha}}$ is a matrix of zeros and $\mathbf{I}_{H_\alpha}$ is a identity matrix. Each embedding is a vector $\boldsymbol{\alpha}^{(s)} \in \mathbb{R}^{H_\alpha}$ representing subject-specific information that remains fixed over time.

\paragraph{Node and community embeddings} Next, let $\boldsymbol{\phi}_n^{(s,\, t)} \in \mathbb{R}^{H_\phi}$ and $\boldsymbol{\psi}_k^{(s,\, t)} \in \mathbb{R}^{H_\psi}$ denote the $n$-th node and the $k$-th community embedding, respectively. To incorporate temporal dynamics, we assume node and community embeddings are related through Markov chains with prior transition distributions $\boldsymbol{\phi}_n^{(s,\, t)} \sim p_{\theta_\phi}(\boldsymbol{\phi}^{(s,\, t)}_n|\boldsymbol{\phi}^{(s,\, t-1)}_n, \, \boldsymbol{\alpha}^{(s)})$ and $\boldsymbol{\psi}_k^{(s,\, t)} \sim p_{\theta_\psi}(\boldsymbol{\psi}^{(s,\, t)}_k|\boldsymbol{\psi}^{(s,\, t-1)}_k, \, \boldsymbol{\alpha}^{(s)})$. We specify each prior to be a normal distribution following
\begin{align}
p_{\theta_\phi}(\boldsymbol{\phi}^{(s,\, t)}_n | \boldsymbol{\phi}^{(s,\, t-1)}_n, \, \boldsymbol{\alpha}^{(s)}) &= \text{Normal}(\boldsymbol{\phi}^{(s,\, t-1)}_n, \, \sigma_\phi\mathbf{I}_{H_{\phi}}) \\ 
p_{\theta_\psi}(\boldsymbol{\psi}^{(s,\, t)}_k | \boldsymbol{\psi}^{(s,\, t-1)}_k, \, \boldsymbol{\alpha}^{(s)}) &= \text{Normal}(\boldsymbol{\psi}^{(s,\, t-1)}_k,\, \sigma_\psi\mathbf{I}_{H_{\psi}})
\end{align}
where the graph embeddings are used for initializing the means, i.e., $\boldsymbol{\phi}_n^{(s,\, 0)}=\boldsymbol{\alpha}^{(s)}$, $\boldsymbol{\psi}_k^{(s,\, 0)}=\boldsymbol{\alpha}^{(s)}$ and the standard deviations $\sigma_\phi, \sigma_\psi \in \mathbb{R}_{>0}$ are hyperparameters controlling how smoothly each embedding changes between consecutive timepoints.



\paragraph{Edge generation} We next describe the edge generative process of a graph snapshot $\mathcal{G}^{(s,\, t)} \in \mathcal{G}^{(1:S,\, 1:T)}$. Similar to~\citet{sun2019vgraph}, for each edge $e^{(s,\, t)}_i = (w^{(s,\, t)}_i,\, c^{(s,\, t)}_i) \in \mathcal{E}^{(s,\, t)}$ we first sample a latent community assignment $z_{i}^{(s,\, t)} \in [1:K]$ from a conditional prior $z_{i}^{(s,\,t)} \sim p_{\theta_z}(z_{i}^{(s,\, t)}|w_i^{(s,\, t)})$ implemented as a categorical distribution
\begin{equation}
\begin{split}
p_{\theta_z}(z_{i}^{(s,\, t)}|w_i^{(s,\, t)}) = \text{Categorical}(\boldsymbol{\pi}_{\theta_z}^{(s,\, t)}), \quad  
\boldsymbol{\pi}_{\theta_z}^{(s,\, t)} = \text{MLP}_{\theta_{z}}(\boldsymbol{\phi}_{w_i}^{(s,\, t)})
\end{split}
\label{eq:prior_source_node}
\end{equation}
where $\text{MLP}_{\theta_{z}}: \mathbb{R}^{H_{\phi}} \rightarrow \mathbb{R}^{K}$ is a $L_z$-layered multilayered perception (MLP) that parameterizes community probabilities using node embeddings indexed by $w^{(s,\, t)}_i$. In other words, each source node $w^{(s,\, t)}_i$ is represented as a mixture of communities. A linked target node $c_{i}^{(s,\, t)} \in [1:N]$ is then sampled from the conditional likelihood $c_{i}^{(s,\, t)} \sim p_{\theta_c}(c_{i}^{(s,\, t)}|z_i^{(s,\, t)})$ which is also implemented as a categorical distribution
\begin{equation}
\begin{split}
p_{\theta_c}(c_{i}^{(s,\, t)}| z_{i}^{(s,\, t)}) = \text{Categorical}(\boldsymbol{\pi}_{\theta_c}^{(s,\, t)}), \quad
\boldsymbol{\pi}_{\theta_c}^{(s,\, t)} = \text{MLP}_{\theta_{c}}(\boldsymbol{\psi}_{z_{i}}^{(s,\, t)})
\label{eq:prior_linked_node}
\end{split}
\end{equation}
where $\text{MLP}_{\theta_{c}}: \mathbb{R}^{H_{\psi}} \rightarrow \mathbb{R}^{N}$ is a $L_c$-layered MLP that parameterizes node probabilities using community embeddings indexed by $z_{i}^{(s,\, t)}$. That is, each community assignment $z_{i}^{(s,\, t)}$ is represented as a mixture of nodes. By integrating out the latent community assignment variable
\begin{equation}
p(c^{(s,\, t)}_{i}|w^{(s,\, t)}_i) = \sum_{z^{(s,\, t)}_{i} \in [1:K]} p_{\theta_c}(c_{i}^{(s,\, t)}| z_{i}^{(s,\, t)}) p_{\theta_z}(z_i^{(s,\, t)}|w_i^{(s,\, t)})
\end{equation}
 we define the likelihood of node $c^{(s,\, t)}_{i}$ being a linked neighbor of node $w^{(s,\, t)}_i$, in a given graph snapshot. 

\paragraph{Factorized generative model} Given this model specification, the joint probability of the observed data and the latent variables can be factorized following
\begin{equation}
\begin{split}
p_\theta(\mathcal{G}^{1:S\, 1:T},\, \Omega^{1:S, 1:T}) ={}& \prod_{s=1}^S \Bigg( p_{\theta_\alpha}(\boldsymbol{\alpha}^{(s)}) \prod_{t=1}^T \Bigg(  \prod_{n=1}^{V}  p_{\theta_\phi}(\boldsymbol{\phi}^{(s,\, t)}_n | \boldsymbol{\phi}^{(s,\, t-1)}_n)  \\ 
& \prod_{k=1}^K p_{\theta_\psi}(\boldsymbol{\psi}^{(s, t)}_k|\boldsymbol{\psi}^{(s, t-1)}_k) \\ 
& \prod_{i=1}^{E^{(s,\, t)}} p_{\theta_z}(z_{i}^{(s,\, t)}|\boldsymbol{\phi}_{w_{i}}^{(s,\, t)})  p_{\theta_c}(c^{(s,\, t)}_{i}|\boldsymbol{\psi}_{z_i}^{(s,\, t)}) \Bigg) \Bigg)  
\label{eq:factorized_generative_model}
\end{split}
\end{equation}
where $\theta = \{\theta_{c}\,, \theta_{z} \}$ is the set of generative model parameters, i.e., NN weights. The generative model of \model summarized in Appendix~\ref{appendix_method}


\subsection{Inference network}
\label{inference_network}


To learn the embeddings, we must infer the posterior distribution over all latent variables conditioned on the observed data $p_\theta(\Omega^{(1:S,\, 1:T)} | \mathcal{G}^{(1:S,\, 1:T)})$. However, exact inference is intractable due the log marginal likelihood
 requiring integrals that are hard to evaluate, i.e., $\log p_{\theta} (\mathcal{G}^{(1:S,\, 1:T)}) = \int_\Omega \log p_{\theta} (\mathcal{G}^{(1:S,\, 1:T)}, \Omega^{(1:S,\, 1:T)})d\Omega$. As a result, we use variational inference~\citep{jordan1999introduction} to approximate the true posterior with a variational distribution $q_\lambda(\Omega^{(1:S, 1:T)})$ with parameters $\lambda$. To do this, we maximize a lower bound on the log marginal likelihood of the DBGs, referred to as the ELBO (\textbf{ev}idence \textbf{l}ower \textbf{bo}und), defined as
\begin{equation}
\mathcal{L}_{\text{ELBO}}(\theta, \lambda) = \mathbb{E}_{q_\lambda} \Bigg[ \log \frac{p_\theta(\mathcal{G}^{1:S,\, 1:T},\, \Omega^{1:S,\, 1:T})}{q_\lambda(\Omega^{(1:S,\, 1:T)})} \Bigg] \leq \log p_\theta(\mathcal{G}^{(1:S,\, 1:T)})
\label{eq:elbo}
\end{equation}
where $\mathbb{E}_{q_\lambda}[ \cdot]$ denotes the expectation taken with respect to the variational distribution $q_\lambda(\Omega^{(1:S,\, 1:T)})$. By maximizing the ELBO with respect to the generative and variational parameters $\theta$ and $\lambda$ we train our generative model and perform Bayesian inference, respectively.

\paragraph{Structured variational distribution} To ensure a good approximation to true posterior, we retain the Markov properties of the node and community embeddings. This results in a structured variational distribution~\citep{hoffman2015structured, saul1995exploiting} which factorizes following
\begin{equation}
\begin{split}
q_\lambda(\Omega^{(1:S,\, 1:T)}) ={}& \prod_{s=1}^S \Bigg( q_{\lambda_\alpha}(\boldsymbol{\alpha}^{(s)}) \prod_{t=1}^T \Bigg( \prod_{n=1}^V q_{\lambda_\phi}(\boldsymbol{\phi}_n^{(s,\,t)}|\, \boldsymbol{\phi}_n^{(s,\,t-1)}) \\ & \prod_{k=1}^K q_{\lambda_\psi}(\boldsymbol{\psi}_k^{(s,\,t)}| \, \boldsymbol{\psi}_k^{(s,\,t-1)}) \prod_{i=1}^{E^{(s,\, t)}} q_{\lambda_z}(z_i^{(s,\,t)}| \, \boldsymbol{\phi}_{w_i}^{(s,\,t)},\, \boldsymbol{\phi}_{c_i}^{(s,\,t)}) \Bigg) \Bigg)
\label{eq:variational_distribution}
\end{split}
\end{equation}
\hspace{-0.5em}
where each distribution is specified to mimic the structure of the generative model so that
\begin{align}
q_{\lambda_\alpha}(\boldsymbol{\alpha}^{(s)}) &= \text{Normal}(\boldsymbol{\mu}^{(s)}_{\lambda_\alpha},\, \boldsymbol{\sigma}_{\lambda_\alpha}^{(s)})  & & 
 \label{eq:alpha_posterior}\\
q_{\lambda_\phi}(\boldsymbol{\phi}_n^{(s,\, t)}| \boldsymbol{\phi}_n^{(s,\, t-1)}) &= \text{Normal}(\boldsymbol{\mu}_{\lambda_\phi}^{(s,\, t)},\, \boldsymbol{\sigma}_{\lambda_\phi}^{(s,\, t)}) &  \{\boldsymbol{\mu}^{(s,\, t)}_{\lambda_\phi},\, \boldsymbol{\sigma}_{\lambda_\phi}^{(s,\, t)}\} &= \text{GRU}_{\lambda_{\phi}}(\boldsymbol{\phi}_n^{(s,\, t-1)})  \\ 
q_{\lambda_\psi}(\boldsymbol{\psi}_k^{(s,\, t)}| \boldsymbol{\psi}_n^{(s,\, t-1)}) &= \text{Normal}(\boldsymbol{\mu}_{\lambda_\psi}^{(s,\, t)},\, \boldsymbol{\sigma}_{\lambda_\psi}^{(s,\, t)}) & \{\boldsymbol{\mu}^{(s,\, t)}_{\lambda_\psi},\, \boldsymbol{\sigma}_{\lambda_\psi}^{(s,\, t)}\} &= \text{GRU}_{\lambda_{\psi}}(\boldsymbol{\psi}_k^{(s,\, t-1)}) \\
q_{\lambda_z}(z_i^{(s,\,t)}| \boldsymbol{\phi}_{w_i}^{(s,\,t)},\, \boldsymbol{\phi}_{c_i}^{(s,\,t)}) & = \text{Categorical}(\boldsymbol{\pi}_{\lambda_z}^{(s,\, t)}) & \boldsymbol{\pi}_{\lambda_z}^{(s,\, t)} &= \text{MLP}_{\lambda_z}( \boldsymbol{\phi}_{w_i}^{(s,\,t)} \odot \boldsymbol{\phi}_{c_i}^{(s,\,t)})
\end{align}

where $\text{GRU}_{\lambda_j}: \mathbb{R}^{H_{j}} \rightarrow \mathbb{R}^{H_{j}}$ is a $L_j$-layered GRU for each $j \in \{\phi, \psi\}$ and $\text{MLP}_{\lambda_z}: \mathbb{R}^{H_{\phi}} \rightarrow \mathbb{R}^{K}$ is $L_z$-layered MLP. Furthermore, we use MLPs to initialize the GRUs with the graph embeddings such that $\boldsymbol{\phi}_n^{(s,\, 0)} = \text{MLP}_{\lambda_\phi}(\boldsymbol{\alpha}^{(s)})$ and $\boldsymbol{\psi}_k^{(s,\, 0)} = \text{MLP}_{\lambda_\psi}(\boldsymbol{\alpha}^{(s)})$ where $\text{MLP}_{\lambda_j}: \mathbb{R}^{N_\alpha} \rightarrow \mathbb{R}^{N_j}$. This allows for subject-specific variation to be incorporated in the temporal dynamics of the node and community embeddings. Another difference with the generative model is now the variational distribution of the community assignment $q_{\lambda_z}(\cdot)$ includes information from neighboring nodes via $c^{(s,\, t)}_i$. Finally, we use the same NN from the generative model to parameterize the variational distribution of the community assignment, i.e., $\lambda_z = \theta_z$. This not only spares additional trainable parameters for
the variational distribution but also further links the variational parameters of $q_\lambda(\cdot)$ to generative parameters
of $p_\theta(\cdot)$ resulting in more robust learning~\citep{farnoosh2021deep}. The set of parameters for the inference network is therefore $\lambda =\{\lambda_\alpha =\{\boldsymbol{\mu}^{(s)}_{\lambda_\alpha},\, \boldsymbol{\sigma}_{\lambda_\alpha}^{(s)} \}_{s=1}^S, \, \lambda_\phi,\, \lambda_\psi,\, \lambda_z=\theta_z\}$.

\paragraph{Training objective} Substituting the variational distribution from \eqref{eq:variational_distribution} and the joint distribution from \eqref{eq:factorized_generative_model} into the ELBO \eqref{eq:elbo} gives the full training objective which can be optimized using stochastic gradient descent. We estimate all gradients using the reparameterization trick~\citep{kingma2013auto} and the Gumbel-softmax trick~\citep{jang2016categorical, maddison2016concrete}. We refer to Appendix \ref{appendix_method} further details on the ELBO and learning the parameters.

%% file: 5-experiments.tex
\section{Experiments}

\begin{table}
\footnotesize
\centering \sisetup{table-number-alignment=center, separate-uncertainty, table-align-uncertainty=false}
\setlength{\tabcolsep}{8.5pt} 
\begin{tabular}{@{}llllll@{}}
\toprule
\multirow{3}{*}{\textbf{Model}} & 
\multicolumn{2}{c}{\textbf{HCP}} &  &
\multicolumn{2}{c}{\textbf{UKB}} \\ 
\cmidrule(lr){2-3} \cmidrule(lr){5-6} & 
\textbf{NLL ($\mathbf{\downarrow}$)} & 
\textbf{MSE ($\mathbf{\downarrow}$)} & &
\textbf{NLL ($\mathbf{\downarrow}$)} & 
\textbf{MSE ($\mathbf{\downarrow}$)}  \\
\midrule
CMN & 
5.999 $\pm$ 0.029 * &
0.050 $\pm$ 0.005 * & &
5.861 $\pm$ 0.017 * & 
0.050 $\pm$ 0.003 * \\
VGAE & 
5.857 $\pm$ 0.017 * &
0.051 $\pm$ 0.002 * & &
5.851 $\pm$ 0.027 * & 
0.061 $\pm$ 0.002  * \\
OSBM & 
5.808 $\pm$ 0.026 * &
0.051 $\pm$ 0.003 * & &
5.726 $\pm$ 0.039 * &
0.052 $\pm$ 0.003 * \\
VGRAPH & 
\underline{5.569 $\pm$ 0.046 }* &
0.022 $\pm$ 0.004 * & &
5.716 $\pm$ 0.037 * & 
0.020 $\pm$ 0.003 * \\
VGRNN & 
5.674 $\pm$ 0.034 * &
\underline{0.011 $\pm$ 0.003} * & &
\underline{5.649 $\pm$ 0.035} * & 
\underline{0.014 $\pm$ 0.002} * \\
ELSM & 
5.924 $\pm$ 0.040 * &
0.081 $\pm$ 0.002 * & &
5.809 $\pm$ 0.024 * & 
0.115 $\pm$ 0.003 * \\
\model & 
\B{4.587 $\pm$ 0.045} &
\B{0.001 $\pm$ 0.002} & &
\B{4.586 $\pm$ 0.084} &
\B{0.004 $\pm$ 0.003} \\ 
\midrule
\midrule

 &
\textbf{AUROC ($\mathbf{\uparrow}$)} & 
\textbf{AP ($\mathbf{\uparrow}$)} &  &
\textbf{AUROC ($\mathbf{\uparrow}$)} & 
\textbf{AP ($\mathbf{\uparrow}$)} \\ 
\midrule
CMN & 
0.665 $\pm$ 0.007 * & 
0.654 $\pm$ 0.006 * & &
0.678 $\pm$ 0.004 * & 
0.668 $\pm$ 0.005 * \\
VGAE & 
0.661 $\pm$ 0.010 * &
0.674 $\pm$ 0.008 * & &
0.688 $\pm$ 0.010 * & 
0.607 $\pm$ 0.009  * \\
OSBM & 
0.655 $\pm$ 0.027 * &
0.675 $\pm$ 0.024 * & &
0.678 $\pm$ 0.032 * & 
0.682 $\pm$ 0.033 * \\
VGRAPH & 
\underline{0.689 $\pm$ 0.004} * & 
0.682 $\pm$ 0.002 * & &
0.664 $\pm$ 0.002 * & 
0.621 $\pm$ 0.001 * \\
VGRNN & 
\underline{0.689 $\pm$ 0.007} * & 
\underline{0.698 $\pm$ 0.006} * & &
\underline{0.698 $\pm$ 0.009} * & 
\underline{0.696 $\pm$ 0.007} * \\
ELSM & 
0.669 $\pm$ 0.004 * & 
0.662 $\pm$ 0.002 * & &
0.661 $\pm$ 0.001 * & 
0.662 $\pm$ 0.002 * \\
\model & 
\B{0.768 $\pm$ 0.026}  &  
\B{0.732 $\pm$ 0.032}  & &
\B{0.786 $\pm$ 0.040}  & 
\B{0.762 $\pm$ 0.038} \\ \bottomrule
\end{tabular}
\caption{Graph reconstruction (top) and dynamic link prediction (bottom) results (mean $\pm$ standard deviation over 5 runs). First and second-best results shown in \textbf{bold} and \underline{underlined}. Statistically significant difference from \model marked *.}
\label{tab:dynamic_link_prediction}
\end{table}

We evaluate \model against baseline models on the tasks of graph reconstruction, dynamic link prediction, and graph classification. Each task is designed to evaluate the usefulness of the learnt embeddings.


\paragraph{Datasets} We construct two multi-subject DBG datasets using publicly available fMRI scans from the Human Connectome Project (HCP)~\cite{van2013wu} and UK Biobank (UKB)~\cite{sudlow2015uk}. We randomly sample $S=300$ subjects ensuring an even male/female split. To create DBGs, we parcellate each scan into $V=360$ region-wise BOLD signals using the Glasser atlas~\cite{glasser2016multi}, apply sliding-window Pearson correlation~\cite{calhoun2014chronnectome} with a non-overlapping window of size and stride of $30$, and threshold the top $5\%$ values of the lower triangle of each correlation matrix as connected following~\citet{kim2021learning}. The described procedure gives $T = 16$ graph snapshots for each subject. Biological sex is taken as graph-level labels. We refer to Appendix~\ref{appendix_datasets} for further details on each dataset.


\paragraph{Baselines} We compare \model against a range of different unsupervised probabilistic baseline models. For static baselines, we include variational graph autoencoder (VGAE)~\citep{kipf2016variational}, a deep generative version of the overlapping stochastic block model (OSBM)~\citep{mehta2019stochastic}, and vGraph (VGRAPH)~\citep{sun2019vgraph}. For dynamic baselines we include variational graph recurrent neural network (VGRNN)~\citep{hajiramezanali2019variational} and evolving latent space model (ELSM)~\citep{gupta2019generative}. For the graph reconstruction and link prediction tasks, we also include a heuristic baseline based on common neighbors between nodes at previous snapshots (CMN). Finally, for graph classification we include a support vector machine which takes as import static FC matrices (FCM)~\citep{abraham2017deriving}. Further details about baseline model can be found in Appendix~\ref{appendix_baselines}.

\paragraph{Implementation} We split both datasets into 80/10/10\% training/validation/test data along the time dimension. We train all models using the Adam optimizer~\cite{kingma2014adam} with decoupled weight decay~\cite{loshchilov2017decoupled}. All baseline hyperparameters are set following their original implementations. For \model, choose the number of communities $K$ based on validation NLL. Finally, we train all models 5 times using different random seeds. Implementation details can be found in Appendix~\ref{appendix_implementation}.

\paragraph{Evaluation metrics} For graph reconstruction, we evaluate the probability of the edges in the test dataset using negative log-likelihood (NLL). We also compare the mean-squared error (MSE) between actual and reconstructed node degree over all test snapshots. For dynamic link prediction, we sample an equal number of positive and negative edges in the test dataset and measure performance using area under the receiver operator curve (AUROC) and average precision (AP). Finally, for graph classification we predict the biological sex for each subjects' DBG and evaluate on accuracy. To predict graph labels, we average node embeddings per subject for the baselines and the community embeddings for \model before training a SVM using 10-fold cross-validation. For comparing models, we use the almost stochastic order (ASO) test~\citep{dror2019deep} with significance level $0.05$ and correct for multiple comparisons~\cite{bonferroni1936teoria}.


\section{Results}

\begin{figure}
\centering
\includegraphics[draft=false, width=\textwidth]{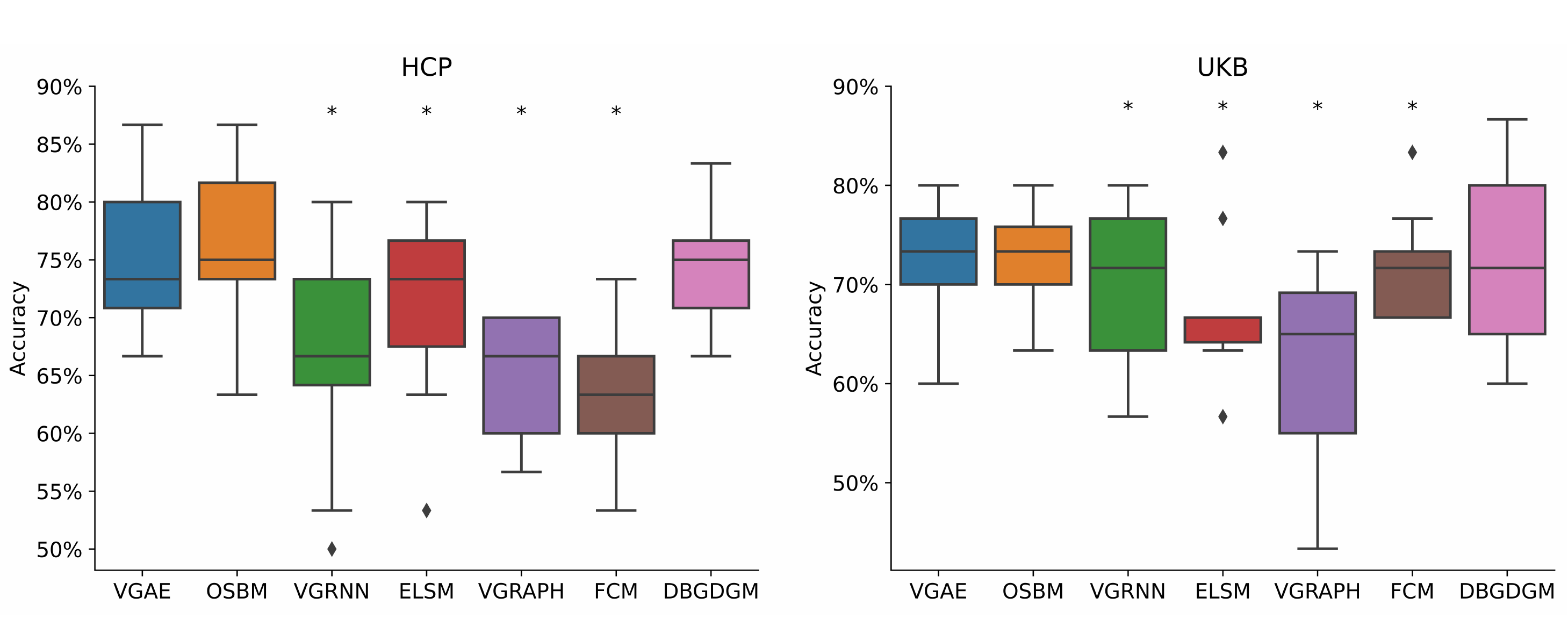}
\caption{\footnotesize Graph classification results (5 runs). Statistical significance from \model marked *.}
\label{fig:gender_classification}
\end{figure}

\paragraph{Dynamic graph reconstruction and link prediction.} We summarize the average test results of all models over 5 runs using optimally tuned hyperparameters.
From \tableref{tab:dynamic_link_prediction}, it is clear that \model outperforms baselines on both tasks. For graph reconstruction, \model shows an $18\%$ and $30\%$ relative improvement in NLL on HCP and UKB, respectively, compared to the second-best baselines. For dynamic link prediction, the relative improvement is $>11\%$ in AUCROC and $>5\%$ in AP compared to second-best baselines depending on dataset. We attribute these statistically significant gains to \model's ability to learn dynamic brain connectivity more effectively. 



\paragraph{Graph classification} 

For graph classification, \model achieves $\sim75\%$ accuracy for HCP and $\sim73\%$ for UKB (see Fig.~\ref{fig:gender_classification}). We outperform 4 baselines and show indiscernible performance to VGAE and OSBM. To show the interpretative power of \model, we re-run the graph classification experiment for HCP with the embeddings of each community separately. We find a community which comprises brain regions in the Cingulo-opercular (CON) and the Somatomotor (SMN) networks, which achieves $68\%$ accuracy. This finding is in agreement with studies that show SMN is predictive of gender \cite{zhang2018functional}.



\paragraph{Interpretability analysis}


\begin{figure}%
\centering
{\includegraphics[width=0.48\textwidth, draft=false]{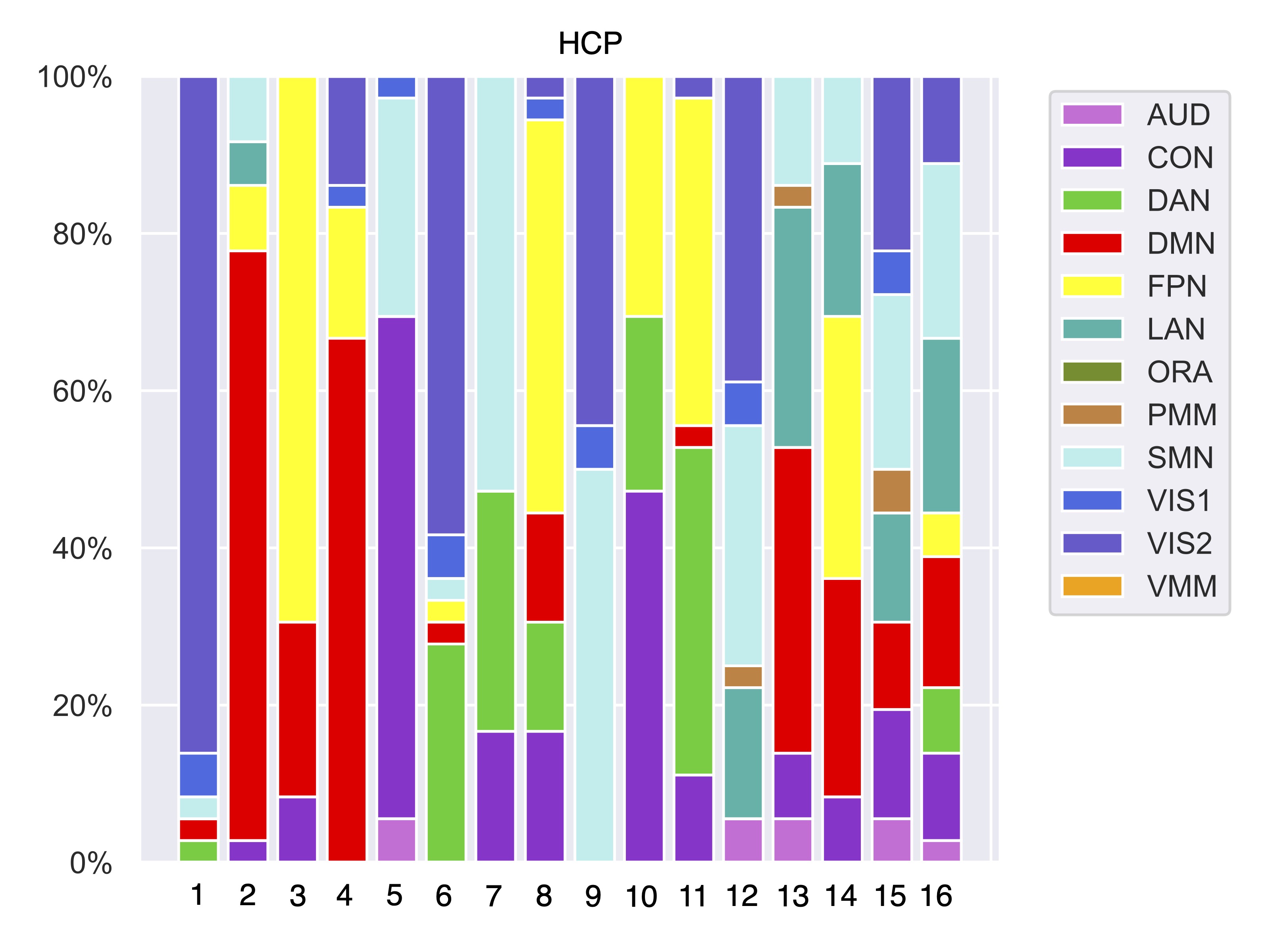}} \hskip1ex
{\includegraphics[width=0.48\textwidth, draft=false]{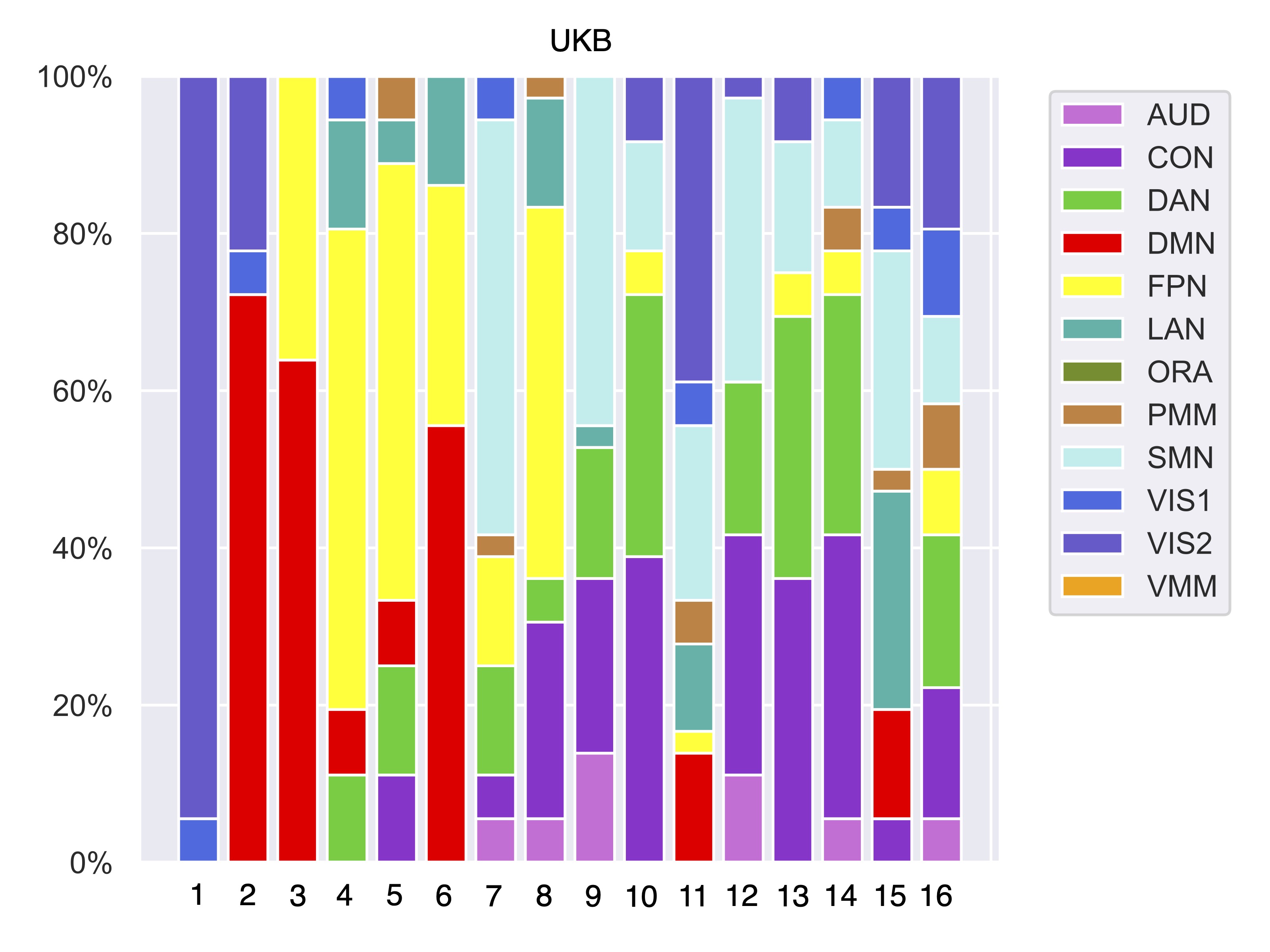}} \hfill
\caption{\footnotesize Overlap between communities learned by \model and FCNs from \citet{ji2019mapping}.}
\label{fig:community}
\end{figure}

We use the learnt distributions over the nodes to calculate overlap between each community and known functional connectivity networks (FCNs) from~\citet{ji2019mapping} (see Appendix \ref{appendix_results}).
~\figureref{fig:community} shows that \model finds communities that significantly overlap with existing FCNs. In particular, nodes in community 1 almost fully corresponds to the visual network (VIS1 + VIS2), which is in keeping with the nature of the experiment (the resting state data was acquired with eyes open and cross-hair fixation). Remarkably, the second and third most homogeneous communities correspond to a large degree to the DMN, which is well known to dominate resting state activity as a whole \cite{yeshurun2021default}. The inspection of additional communities and respective predictive power, along with their evolution in time at the region-of-interest granularity, has the potential to unveil the yet largely unexplored relationships between dynamic brain connectivity changes and, e.g. psychiatric or neurological disorders \cite{heitmann2017putting}.










%% file: 6-conclusion.tex
\section{Conclusion}
We propose \model, a hierarchical DGM designed for unsupervised representing learning of DBGs. Specifically, \model jointly learns  graph-, community-, and node-level embeddings that outperform baselines on classification, interpretability, and dynamic link prediction with statistical significance. Moreover, an analysis of the learnt dynamic community-node distributions shows significant overlap with existing FCNs from neuroscience literature further validating our method.









%% file: 7-acknowledgments.tex
\midlacknowledgments{This work is supported by The Alan Turing Institute under the EPSRC grant EP/N510129/1. Data were provided [in part] by the Human Connectome Project, WU-Minn Consortium (Principal Investigators: David Van Essen and Kamil Ugurbil; 1U54MH091657) funded by the 16 NIH Institutes and Centers that support the NIH Blueprint for Neuroscience Research; and by the McDonnell Center for Systems Neuroscience at Washington University.}

%% file: 9-appendix.tex
\appendix



\section{Related work}
\label{appendix_related_work}

\paragraph{Dynamic graph generative models} Classic generative models for graph-structured data are designed for capturing a small set of specific properties (e.g., degree distribution, eigenvalues, modularity) of static graphs~\citep{erdos1960evolution, barabasi1999emergence, nowicki2001estimation}. DGMs that exploit the learning capacity of NNs are able to learn more expressive graph distributions~\citep{mehta2019stochastic, kipf2016variational, sarkar2020graph}. Recent DGMs for dynamic graphs are majority VAE-based~\citep{kingma2013auto} and cannot learn community representations~\citep{hajiramezanali2019variational, gracious2021neural, zhang2021disentangled}. The few that do, are designed for static graphs~\cite{sun2019vgraph, khan2021unsupervised, cavallari2017learning}. 

\paragraph{Learning representations of dynamic brain graphs} Unsupervised representation learning methods for DBGs tend to focus on clustering DBGs into a finite number of connectivity patterns that recur over time~\citep{allen2014tracking, spencer2022using}. Community detection is another commonly used method but mainly applied to static brain graphs~\citep{pavlovic2020multi, esfahlani2021modularity}. Extensions to DBGs are typically not end-to-end trainable and do not scale to multi-subject datasets~\citep{ting2020detecting, martinet2020robust}. Recent deep learning-based methods are predominately GNN-based~\citep{kim2021learning, dahan2021improving}. Unlike \model, these methods are supervised and focus on learning deterministic node- and graph-level representations.

\section{Method}
\label{appendix_method}
\subsection{Generative model}
Algorithm~\ref{alg:generative_model} summarizes the generative model for \model.
\begin{algorithm2e}[!ht]
\caption{\model generative model}
\KwIn{$\{ \mathcal{E}^{(s,\, t)} \}_{s,\, t =1}^{S,\, T}$}
\Parameter{$K$, $H_\alpha$, $H_\psi$, $H_\phi$, $L_\psi$, $L_\phi$, $L_z$, $\sigma_\psi^2$, $\sigma_\phi^2$,}
\BlankLine
\kwInit{$\mathcal{D} \gets \emptyset$} 
\For{$s \gets 1$ \KwTo $S$}{
$\boldsymbol{\alpha}^{(s)} \sim p(\boldsymbol{\alpha}^{(s)}) = \text{Normal}(\mathbf{0}_{H_\alpha}, \mathbf{I}_{H_\alpha})$ \\
\For{$t \gets 1$ \KwTo $T$}{
\For{$k \gets 1$ \KwTo $K$}{
$\boldsymbol{\psi}_k^{(s, t)} \sim p(\boldsymbol{\psi}^{(s,\, t)}_k 
| \boldsymbol{\psi}^{(s,\, t-1)}_k
) 
= \text{Normal}(\boldsymbol{\psi}^{(s,\, t-1)}_k,\, \sigma_\psi\mathbf{I}_{H_{\psi}})$}

\For{$n \gets 1$ \KwTo $V$}{
$\boldsymbol{\phi}_n^{(s, t)} \sim p(\boldsymbol{\phi}^{(s,\, t)}_n 
| \boldsymbol{\phi}^{(s,\, t-1)}_n
) 
= \text{Normal}(\boldsymbol{\phi}^{(s,\, t-1)}_k,\, \sigma_\phi\mathbf{I}_{H_{\phi}})$}
$\Tilde{\mathcal{E}}^{(s,\,t)} \gets \emptyset$ \\
\For{$i \gets 1$ \KwTo $|\mathcal{E}^{(s,\,t)}|$}{
$z_{i}^{(s,\, t)} \sim p(z_{i}^{(s,\, t)}|w_i^{(s,\, t)}) = \text{Categorical}(f_{\theta_{\pi}}(\boldsymbol{\phi}_{w_i}^{(s,\, t)}))$ \\
$c_{i}^{(s,\, t)} \sim p(c_{i}^{(s,\, t)}|z_{i}^{(s,\, t)}) = \text{Categorical}(f_{\theta_{\pi}}(\boldsymbol{\psi}_{z_i}^{(s,\, t)}))$ \\
$\Tilde{\mathcal{E}}^{(s,\,t)}  \gets \Tilde{\mathcal{E}}^{(s,\,t)} \cup \{(w_i^{(s, \, t)}, c_i^{(s, \, t)}) \}$  \\}
$\mathcal{G}^{(s,\, t)} \gets (\mathcal{V},\, \Tilde{\mathcal{E}}^{(s,\, t)})$ \\
 $\mathcal{D}  \gets \mathcal{D} \cup \{ \mathcal{G}^{(s,\, t)} \}$
 }}
\label{alg:generative_model}
\end{algorithm2e}

\subsection{Training objective and learning the parameters}
Substituting the variational distribution from \eqref{eq:variational_distribution} and the joint distribution from \eqref{eq:factorized_generative_model} into the ELBO \eqref{eq:elbo} gives the full training objective defined as
\begin{align}
\mathcal{L}_{\text{ELBO}}(\theta, \lambda) ={}& \sum_{s=1}^S \sum_{t=1}^T \sum_{i=1}^{E^{(s,\, t)}} \Bigg( \mathbb{E}_{q_{\lambda_z}q_{\lambda_\psi}}\Big[\log p_\theta(c_{i}^{(s,\, t)}|w_{i}^{(s,\, t)},\, \boldsymbol{\psi}^{(s,\, t)}_{z_{i}})\Big] \notag \\
&-\mathbb{E}_{q_{\lambda_{\phi}}}\Big[\text{D}_{\text{KL}}[q_{\lambda_z}(z_i^{(s,\,t)}| \, \boldsymbol{\phi}_{w_i}^{(s,\,t)},\, \boldsymbol{\phi}_{c_i}^{(s,\,t)}) || p_{\theta_z}(z_i^{(s,\,t)}| \, \boldsymbol{\phi}_{w_i}^{(s,\,t)})]\Big] \Big) \notag \\
&-  \sum_{s=1}^S \Bigg( \text{D}_{\text{KL}}[q_{\lambda_\alpha}(\boldsymbol{\alpha}^{(s)}) || p_{\theta_\alpha}(\boldsymbol{\alpha}^{(s)})] \sum_{t=1}^T \Bigg(  \label{eq:elbo_reconstruction} \\
&- \sum_{n=1}^V  \mathbb{E}_{q_{\lambda_\phi}}\Big[\text{D}_{\text{KL}}[q_{\lambda_\phi}(\boldsymbol{\phi}_n^{(s,\,t)}| \, \boldsymbol{\phi}_n^{(s,\,t-1)}) || p_{\theta_\phi}(\boldsymbol{\phi}_n^{(s,\,t)}| \, \boldsymbol{\phi}_n^{(s,\,t-1)})]\Big] \notag  \\
&- \sum_{k=1}^K \mathbb{E}_{q_{\lambda_\psi}}\Big[\text{D}_{\text{KL}}[q_{\lambda_\psi}(\boldsymbol{\psi}_k^{(s,\,t)}| \, \boldsymbol{\psi}_k^{(s,\,t-1)}) || p_{\theta_\psi}(\boldsymbol{\psi}_k^{(s,\,t)}| \, \boldsymbol{\psi}_k^{(s,\,t-1)})]\Big] \Bigg) \Bigg) \notag
\end{align}
where $\text{D}_{\text{KL}}[\cdot||\cdot]$ denotes the Kullback-Leibler (KL) divergence. By maximizing \eqref{eq:elbo_reconstruction}, the parameters $(\theta, \lambda)$ of the generative
model and inference network can be jointly learnt. 


\paragraph{Learning the parameters}  In order to use efficient stochastic gradient-based optimization techniques~\citep{robbins1951stochastic} for learning $(\theta, \, \lambda)$, the gradient of the ELBO has to be estimated. The main challenge of this is obtaining gradients of the variables under expectation, i.e., $\mathbb{E}_{q_*}[\cdot]$, since they are sampled. To allow gradients to flow through these sampling steps, we use the reparameterization trick~\citep{kingma2013auto, rezende2014stochastic} for the normal distributions and the Gumbel-softmax trick~\citep{jang2016categorical, maddison2016concrete} for the categorical distributions. All gradients are now easily computed via back-propagation~\citep{rumelhart1986learning} making \model end-to-end trainable. In addition, we analytically
calculate the KL terms for both normal and categorical distributions, which
leads to lower variance gradient estimates and faster training as compared to noisy Monte Carlo estimates.

\section{Datasets}
\label{appendix_datasets}

To create multi-subject DBG datasets, we use real fMRI scans from the UK Biobank~\cite{sudlow2015uk} and Human Connectome Project~\cite{van2013wu}. Both data sources represent well-characterized population cohorts that have undergone standardized neuroimaging and clinical assessments to ensure high quality.

\paragraph[]{UK Biobank\footnote{\url{https://www.ukbiobank.ac.uk}} (UKB)} The UKB dataset consists of $S=300$ resting-rate fMRI scans (i.e. 3D image of the brain taken over consecutive timepoints) randomly sampled from the v1.3 January 2017 release ensuring an equal male/female split (i.e. sex balanced) with an age range of $44-57$ years. The total number of images for each scan is $490$ timepoints (6 minutes duration with a repetition time of 0.74s). The dataset is minimally preprocessed following the pipeline described in \citet{alfaro2018image}.

\paragraph[]{Human Connectome Project\footnote{\url{https://www.humanconnectome.org}} (HCP)} The HCP dataset similarly consists of $S=300$ sex balanced resting-state fMRI scans randomly sampled from the S1200 release with an age range of $22-35$ years. Only images from the first scanning-session using left-right phase encoding are used. The total number of images for each scan is $1,200$ timepoints (15 minutes duration with a repetition time of 0.72s). The dataset is minimally preprocessed following the pipeline described in \citet{glasser2013minimal}

 \paragraph[]{Further preprocessing} The fMRI scans from each dataset are further preprocessed to create DBGs. Firstly, each scan is transformed into a multivariate timeseries of BOLD signals using the Glasser atlas~\cite{glasser2016multi} to average voxels within $V=360$ brain regions. Next, to ensure comparability with UKB, we truncate the length of HCP timeseries to $490$ timepoints. Following the commonly used sliding-window method~\cite{calhoun2014chronnectome}, we use Pearson correlation to calculate FC matrices within non-overlapping windows of length $1 < W \leq 490$ along the temporal dimension. At every window, we create an edge set of a unweighted and undirected graph with no self-edges by thresholding the top $1 \leq \epsilon < 100$ percentile values of the lower triangle of the FC matrix (excluding the principal diagonal) as connected following \citet{kim2021learning}. For both datasets, we choose $W=30$ and $\epsilon=5$ resulting in $T=\lfloor 490/30 \rfloor = 16$ graph snapshots each with $E^{(s,\, t)}=\lfloor(360(360-1)/2)(5/100)\rfloor=3,231$ edges. 


\section{Baselines}
\label{appendix_baselines}

We compare \model against a range of static and dynamic unsupervised graph representation learning baseline models, all with publicly available code. In particular, we focus on baselines that are generative and can quantify uncertainty. We leave comparisons to popular deterministic baselines such as DynamicTriad~\cite{zhou2018dynamic}, DySAT~\cite{sankar2020dysat}, and DynNode2Vec~\cite{mahdavi2018dynnode2vec} for future work. Furthermore, since all of the baselines were originally designed to model large single-graph datasets, we had to adapt each implementation to work with smaller multi-graph datasets.

\paragraph[]{Variational graph auto encoder\footnote{\url{https://github.com/tkipf/gae}} (VGAE){\normalfont~\cite{kipf2016variational}}} An extension of the variational autoencoder~\citep{kingma2013auto} (VAE) for graph structured data. Specifically, VGAE uses a graph convolutional network (GCN)~\cite{kipf2016semi} to learn a distribution over node embeddings. Originally designed for static graphs, we train VGAE on each dynamic graph snapshot independently.

\paragraph[]{Overlapping stochastic block model\footnote{\url{https://github.com/nikhil-dce/SBM-meet-GNN}} (OSBM){\normalfont~\cite{mehta2019stochastic}}} A deep generative version of the overlapping stochastic block model~\citep{miller2009nonparametric}. In particular, OSBM places a stick-breaking prior over the number of communities which allows the model to automatically infer the optimal number of communities from the data during training. Similar to VGAE, OSBM uses a GCN to parameterize the distribution over node embeddings and is designed for static graphs.

\paragraph[]{Variational graph RNN\footnote{\url{https://github.com/VGraphRNN/VGRNN}} (VGRNN){\normalfont~\cite{hajiramezanali2019variational}}} An extension of VGAE for dynamic graphs. Using a modified graph RNN architecture, VGRNN is able to learn dependencies between and within changing graph topology over time. Similar to \model, the prior distribution over node embeddings is parameterized using hidden states from previous timepoints.

\paragraph[]{Evolving latent space model\footnote{\url{https://github.com/sh-gupta/ELSM}} (ELSM){\normalfont~\cite{gupta2019generative}}} A generative model for dynamic graphs that learns node embeddings and performs community detection. In particular, node embeddings are initially sampled from a Gaussian mixture model over communities and then evolved over time using an LSTM. Unlike the previous baselines, ELSM does not use a GNNs to parameterize model distributions.

\paragraph[]{vGraph\footnote{\url{https://github.com/fanyun-sun/vGraph}} (VGRAPH){\normalfont~\cite{sun2019vgraph}}} Similar to \model, VGRAPH simultaneously learns node embeddings and community assignments by modeling nodes as being generated from a mixture of communities. The generative process of VGRAPH also relies on edge information. Since VGRAPH only models static graphs, we train it on each dynamic graph snapshot independently.

\paragraph[]{Common neighbors (CMN)} In light of recent work demonstrating that heuristic methods are able to outperform deep-learning based models on dynamic link prediction tasks~\citep{skarding2022robust, poursafaei2022towards}, we include our own heuristic-based generative model baseline. More formally, let $\boldsymbol{\pi}_{v_i}^{(t)} \in [0, 1]^V$ denote a vector of Jaccard index scores for node $v_i^{(t)} \in \mathcal{V}$ with all other nodes $v^{(t)}_j \in \mathcal{V}$ for $i \neq j$. The Jaccard index between two nodes $v_i^{(t)},\, v^{(t)}_j  \in \mathcal{V}$ is defined $|\Gamma(v_i^{(t)}) \cap \Gamma(v^{(t)}_j)| /  |\Gamma(v_i^{(t)}) \cup \Gamma(v^{(t)}_j)|$ where $\Gamma(v_i^{(t)})$ denotes the set of neighbors of node $v_i^{(t)}$. We define the probability of node $v^{(t)}_i$ having a linked neighbor $v_j^{(t)}$ at snapshot $t$ as
\begin{equation}
p(v_j^{(t)} | v_i^{(t)}) = \text{Categorical}(\boldsymbol{\pi}_{v_i}^{(t-1)}).
\end{equation}
This simple generative model captures the intuition that nodes are more likely to form links if they had common neighbors in a previous snapshot. 

\section{Implementation details}
\label{appendix_implementation}

\paragraph{Software and hardware} All models are developed in Python 3.7~\citep{python37} using scikit-learn 1.1.1~\citep{pedregosa2011scikit}, PyTorch\citep{paszke2019pytorch}, and numpy 1.1.1~\citep{numpy}. Statistical significance tests are carried out using deep-significance 1.1.1~\citep{ulmer2022deep}. Experiments are performed on a Linux server (Debian 5.10.113-1) with a NVIDIA RTX A6000 GPU with 48 GB memory and 16 CPUs.  

\paragraph{Training and testing} All baselines are implemented as per the original paper and/or code repository given in Appendix~\ref{appendix_baselines}. For the static graph baselines VGAE, OSBM, VGRAPH we train on each snapshot independently and use the node and/or community embeddings at the last training snapshot to make predictions.

\begin{figure}[!htbp]
\centering
{\includegraphics[width=0.48\textwidth, draft=False]{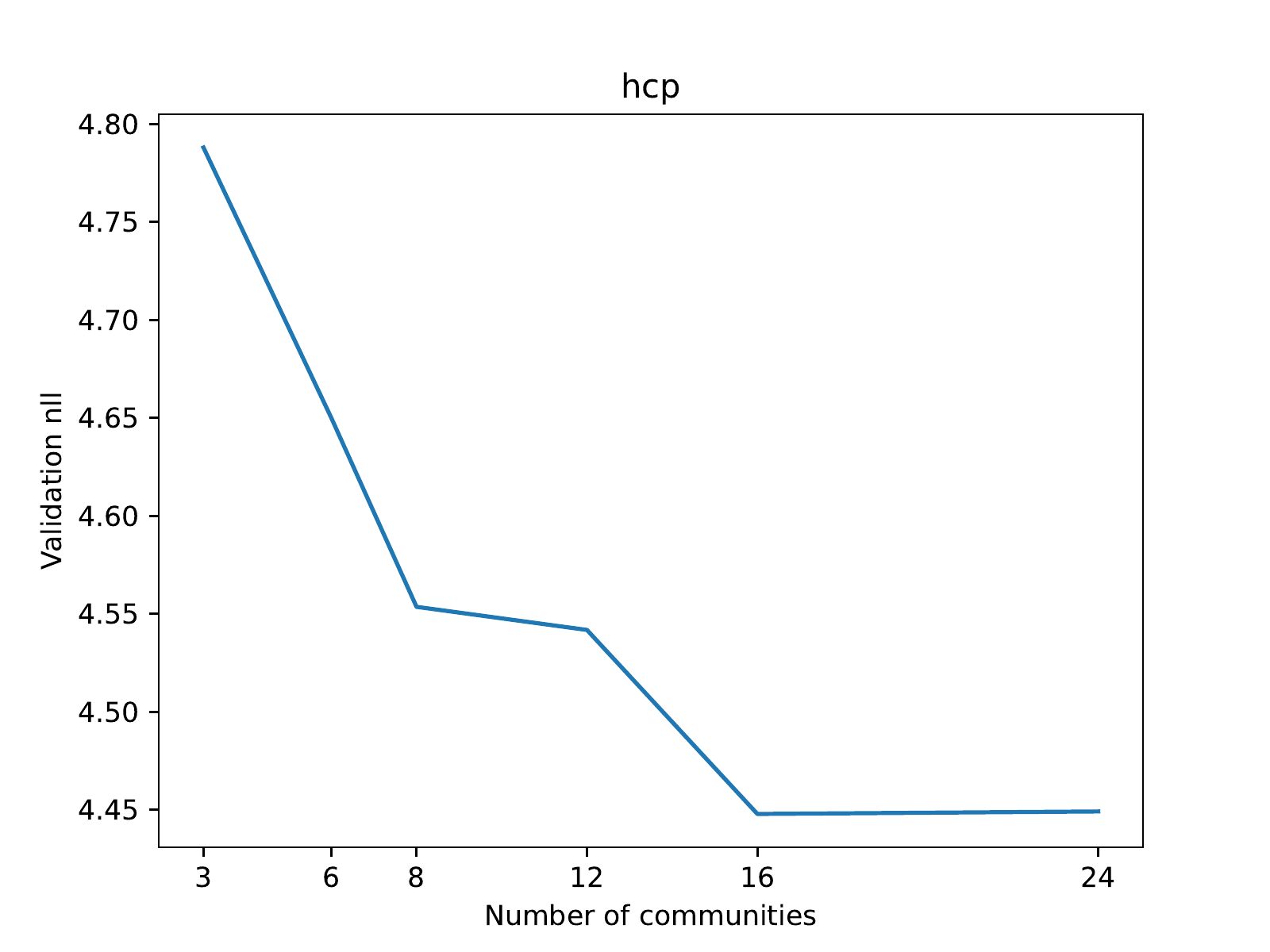}} \hskip1ex
{\includegraphics[width=0.48\textwidth, draft=False]{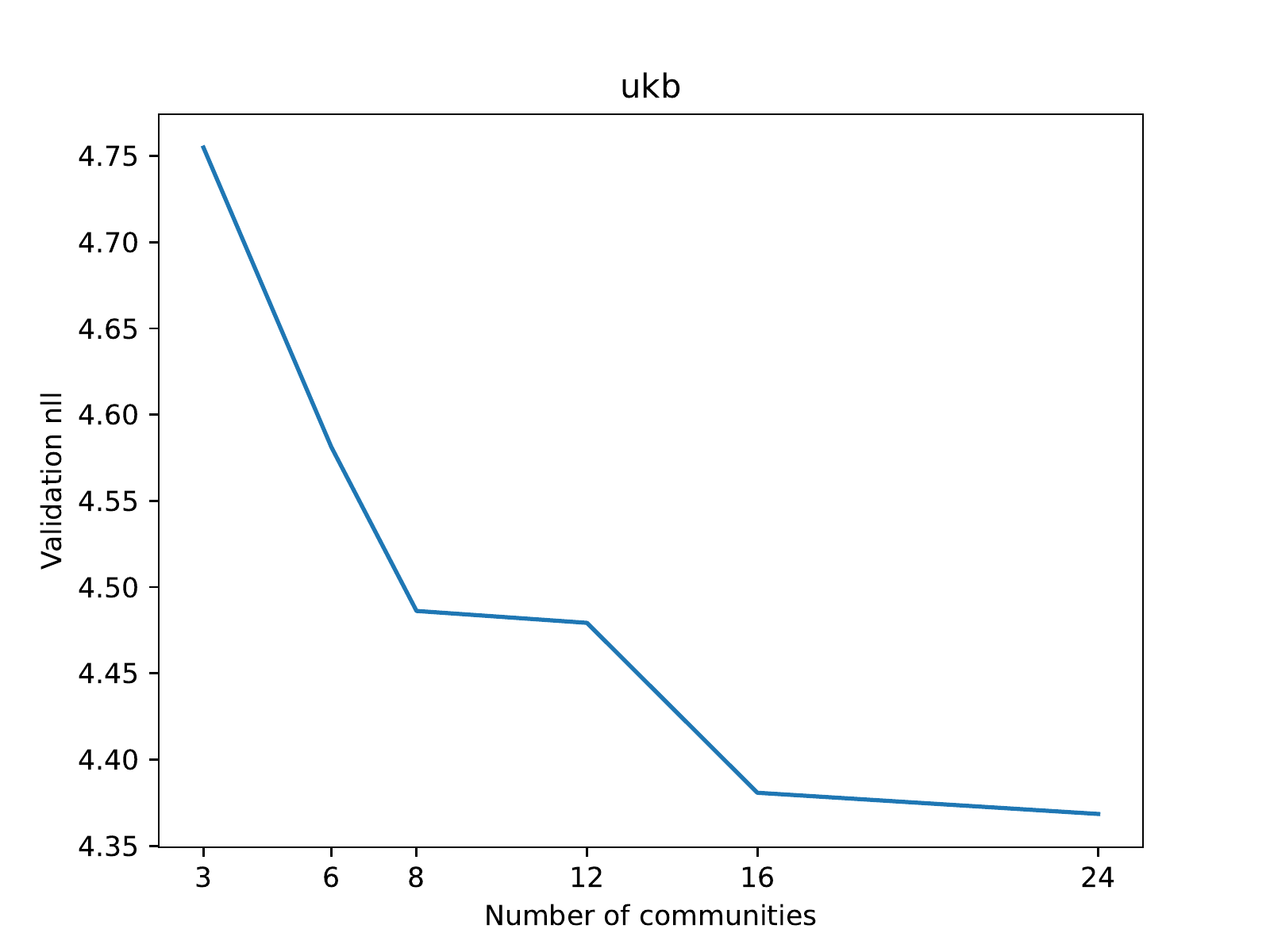}} \hfill
\caption{Elbow plot for finding the optimal number of communities $K$.}
\label{fig:elbow_plot}
\end{figure}

\paragraph{Hyperparameter optimization} We use model and training hyperparameter values described in the original implementation of each baseline as a starting point for tuning on the validation dataset. Since searching for optional values for each hyperparameter configuration was outside the scope of this paper, we focus mainly on tuning the dimensions of hidden layers. For \model, we use a learning rate of 1$e$-4 with a weight decay of $0$. We choose the number of communities $K \in \{3, 6, 8, 12, 16, 24\}$ based on lowest average validation NLL (see \figureref{fig:elbow_plot}). In the generative model, we fix the temporal smoothness hyperparameters $\sigma_\phi = \sigma_\psi = 0.01$. In the inference network, we fix the number of layers for all NNs to $\ L_\phi = L_\psi = L_z = 1$. For the Gumbel-softmax reparameterization trick we anneal the softmax temperature parameter starting from a maximum of 1 to a minimum of 0.05 at a rate of 3$e$-4. Finally, we train all models for $1,000$ epochs using early-stopping with a patience of $15$ based on the lowest validation NLL. 

\section{Interpretability analysis}
\label{appendix_results}

Using \model, for each community we average the node distributions across subjects and timepoints and take the top $10\%$ most probable nodes. We use these high probability nodes to calculate overlap between each community and the brain regions that comprise each functional network from~\citet{ji2019mapping}. More specifically, the coloured proportions in \figureref{fig:community} represent the proportion of top nodes in each community, which belong to a given functional network.

\begin{table}[!htbp]
\centering  
\begin{tabular}{@{}ll@{}}
\toprule
\textbf{Abbreviation} & 
\textbf{Functional network} \\ 
\midrule
 AUD & Auditory network\\
 CON & Cingulo-opercular network\\
 DAN & Dorsal-attention network\\
 DMN & Default mode network\\
 FPN & Frontoparietal network\\
 LAN & Language network\\
 ORA & Orbito-affective network\\
 PMM & Posterior-multimodal network\\
 SMN & Somatomotor network\\
 VIS1 & Visual network 1 \\
 VIS2 & Visual network 2 \\
 VMM & Ventral-multimodal network\\
 \bottomrule
\end{tabular}
\caption{Functional connectivity networks (FCNs) from \citet{ji2019mapping}}
\label{tab:network_notation}
\end{table} 